\documentclass[conference]{IEEEtran}\IEEEoverridecommandlockouts
\usepackage{cite}
\usepackage{amsmath,amssymb,amsfonts}
\usepackage{algorithmic}
\usepackage{graphicx}
\usepackage{textcomp}
\usepackage{xcolor}
\usepackage{epsfig}
\usepackage{float}
\usepackage{bm}
\usepackage{graphicx}
\usepackage{amsfonts,amssymb}
\usepackage{booktabs}
\usepackage{array}
\usepackage{tabularx}
\usepackage{multirow}
\usepackage{subfigure}
\usepackage{color}
\usepackage{epstopdf}
\usepackage{makecell}
\usepackage{url}
\usepackage{gensymb}
\usepackage{amsmath}
\usepackage{xcolor}
\usepackage[implicit=false]{hyperref}

\usepackage[ruled]{algorithm2e}
\def\BibTeX{{\rm B\kern-.05em{\sc i\kern-.025em b}\kern-.08em
		T\kern-.1667em\lower.7ex\hbox{E}\kern-.125emX}}

\begin{document}
	
\title{A Two-stage Method for Non-extreme Value Salt-and-Pepper Noise Removal\vspace{-.5em}}

\author{\IEEEauthorblockN{Renwei Yang$^{\ast}$, YiKe Liu$^{\ast}$, Bing Zeng}
	\IEEEauthorblockA{School of Information and Communication Engineering\\
		University of Electronic Science and Technology of China, Chengdu, China\\}
	\vspace{-1.5em}
	\thanks{*indicates equal contributions}
}
\maketitle

\begin{abstract}
	There are several previous methods based on neural network can have great performance in denoising salt and pepper noise. However, those methods are based on a hypothesis that the value of salt and pepper noise is exactly 0 and 255. It is not true in the real world. The result of those methods deviate sharply when the value is different from 0 and 255. To overcome this weakness, our method aims at designing a convolutional neural network to detect the noise pixels in a wider range of value and then a filter is used to modify pixel value to 0, which is beneficial for further filtering. Additionally, another convolutional neural network is used to conduct the denoising and restoration work.
\end{abstract}

\vspace{.5em}

%\vspace{0.5em}
\section{Introduction}
	
Salt and pepper noise is a kind of common noise in images, it is a kind of white or black spots that appear randomly. Salt and pepper noise is also called impulse noise. Salt and pepper noise may be caused by sudden strong interference on the image signal, analog-to-digital converter, or bit transmission errors. For instance, a failed sensor causes the pixel value to be the minimum value, and a saturated sensor causes the pixel value to be the maximum value. Single image denoising is an important image restoration method which can improve the quality of image with low cost. This project aimed at designing and training a group of neural networks to do the single image denoising salt and pepper noise work. 

Single image denoising is a traditional low-level task in computer vision field. Several previous methods have achieved some great results. In 2017, He et.al \cite{b1} proposed ResNet which is a very deep convolutional neural network based on residual study. This structure of network reach great results in recognition tasks and have great influence on designing of deep neural network in many other fields. Zhang et.al \cite{b2} used a deep convolutional neural network to study the residual part of noise image which is the noise. This method overcome one weakness of deep neural network that it is easy to loss the feature of original image and reached high PSNR in denoising both additive white gaussian noise and salt and pepper noise. However, there are still some noise pixels in the smooth area of the denoised image. Ronneberger et.al\cite{b3} used a network of encoding and decoding structure to do the denoising work. Networks of this kind of structure are widely used in image restoration tasks and have good performance.

However, previous work on using neural network for denoising salt and pepper noise are based on a hypothesis that the value of salt and pepper noise is exactly 0 and 255 which is not true in the real life. When the value of noise spot changes a little in the noised- pictures, the PSNR of the processed picture decrease sharply. 

Our project is aimed at solving this problem. In this project, a two-stage method is proposed. In the first stage, all salt and pepper values are converted to 0 by a forward deep neural network. In the second stage, a DRUnet is applied to remove these black pixels. The main idea of the first step is to design a convolutional neural network for detecting the position of noise pixels and separate it from the clean part of image. Then a filter is added to transfer all the value of noise pixels to 0. Thus, the processed image can be rewritten as,
\begin{equation}
	y^\prime=\ y-\ e =
	\begin{cases}
		&x, \text{ clean pixel y}\\
		&0, \text{ noise pixel y} 
	\end{cases}
	\tag{1}
\end{equation}

for y is the original corrupted image pixel,  \(y^\prime\) is the processed corrupted image pixel, x is the clean image pixel and e is the error. The purpose of converting noise value to 0 is based on the empirical experiment result that, the DRUnet can deal with noise well, if the noise is of one fixed value.

Finally, the DRUnet is utilized in the second stage to denoise. This DRUnet is Unet which is inserted with residual blocks, dawn sampling blocks, up sampling blocks and restoration blocks is used to de the denoising work. What needs to be emphasized is that the Unet have serval direct connection through the network for better performance.

The contribution of this work is listed as follows:
\begin{itemize}
	\item A two-stage method is proposed, which is aimed at removing non-extreme value salt and pepper noise.
	\item In the training step of forward deep neural network , penalty factor is utilized adjust the missing alarm rate and false alarm rate.
	\item The performance indicates that this method can enhance the quality visually and quantitatively.
\end{itemize}
	
\section{Related works}

\subsection{Optimizer}

There are several available optimizers provided in the pytorch package. The optimize process can be shown by a mathematical formula as: 
\begin{equation}
	p^\prime=p-\widehat{g_t}\ast lr
	\tag{2}
\end{equation}
p is the original parameter in the model, \(p^\prime\) is the updated parameter in the model, lr is the learning rate which decides the step size of optimizer, \(\widehat{g_t}=F(g_t)\), \(g_tis\) the gradient of the model, F is the formula of optimizer. SGD optimizer\cite{b4} is the short of stochastic gradient descent, and this method stochastically selects training samples from the training set to update the gradients. There are several parameters in this optimizer as learning rate, momentum, dampening, weight decay and nesterov. Mathematical formulas below will describe this optimizer:
\begin{align}
	v_t&= v_{t-1}\ast mementum+\ g_t\ast\left(1-dampening\right)
	\tag{3a}\\
	g_t&= g_t+p\ast weight_decay 
	\tag{3b}\\
	v_t &= g_t+ v_t\ast momentum
	\tag{3c}
\end{align}
This optimizer has advantages as fast speed when using mini batch, alleviate the problem of no momentum to a certain extent. However, it faces the problem that it may stuck in the local optimization. 

Another widely used optimizer Adam\cite{b5} adds gradient moving average and deviation correction for better performance. This optimizer can be shown as below mathematical formulas:
\begin{align}
	m_{t\ } &=\ \beta_1\ast m_{t-1}+\left(1\ -\beta_1\right)\ast g_t
	\tag{4a}\\
	v_t &=\ \ \beta_2\ast v_{t-1}+\left(1\ -\beta_2\right)\ast(g_t)^2
	\tag{4b}\\
	\widehat{m_{t\ }} &=\ \frac{m_{t\ }}{1-{(\beta_1)}^t}
	\tag{4c}\\
	\widehat{v_{t\ }} &=\ \frac{v_{t\ }}{1-{(\beta_2)}^t}
	\tag{4d}\\
	\theta_t &=\theta_{t-1}-\frac{\widehat{m_{t\ }}}{\sqrt{\widehat{v_{t\ }}}+\in}lr
	\tag{4e}
\end{align}
\( \in\) is a small parameter added to the denominator to avoid division 0.

Both of this two widely used optimizer are used in our project to compare and get better performance.

\subsection{Residual learning}
Residual learning of CNN\cite{b2}was first used in order to solve performance degradation problem when the depth of network increasing. Residual network learns a residual mapping based on the assumption that the residual mapping is much easier to be learned than the original unreferenced mapping. Deep CNN network with a residual learning strategy can be easily trained and improved accuracy in tasks of image classification and object detection. Meanwhile, residual units are also used in some low-level vision problems such as image denoising\cite{b1}, single image super-resolution \cite{b6} and color image demosaicking\cite{b7}.

The proposed of our first step is to use a deep CNN network to study the residual part between the clean image and the noised image and then modify the residual part to fit the noise map which shows the distribution of all the noise pixels.

\subsection{Unet}
In FCN \cite{b8}, deep learning has been firstly utilized in image segmentation. As the improved version of FCN, the concept of Unet was firstly proposed by Ronneberger et.al \cite{b3}, aimed at medical iamge segmentation. This structure shapes like letter U, thus it is named so. The first half part of Unet is down sampling, the last half is up sampling, and skip connection is applied between layers with same size, to enhance the performance. DRUnet is a modified version of Unet, proposed by Zhang et.al \cite{b9}.In our method, DRU net is used to remove pepper noise, but the input of DRU net is reduced from two channels to one channel. In the original method, the input is a Noise Levele Map and the Noisy Image, and only the latter is remained in our method. DRUnet is proposed to handle several low-level vision image restoration tasks, such as denoising, inpainting, super-resolution. The structure used in this paper is shown in Figure 1.
\begin{figure}[htbp]
	\centering
	\includegraphics[width=9cm]{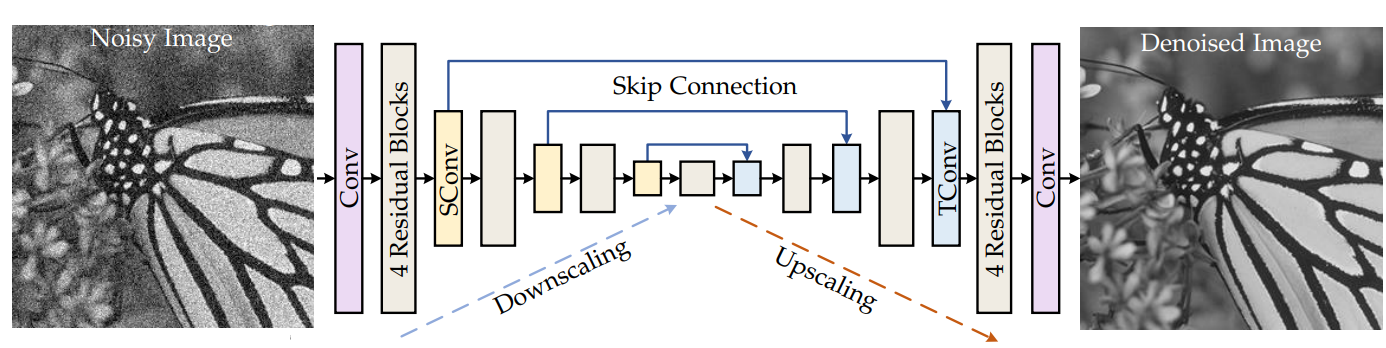}
	\caption{The architecure of the network used in our method}
\end{figure}

In the figure, it is indicated that the downscaling and upscaling parts include four kinds of layers. The purple layer is the normal convolution layer, followed by 4 residual blocks, represented by gray layer. Then, the yellow layer indicates the strided convolution. It is used to down sample the feature map with half height and width, and double the feature map channels. In contrast, the blue layer is the transposed convolution, which does the opposite procedure to recovery the image. Each pair of strided and transposed convolution is skip connected, allowing the feature to propagate directly to the deep layers. 

It is worth noting that, both the residual block \cite{b1} and the whole Unet have utilized skip connection. There is slice difference between their formation and purpose. In ResNet, the shortcuts is realized by element-wise addition, and it is mainly used for preventing gradient vanish. Since the gradient can be back propagated directly from deep layer to shallow layer. In Unet, the skip connection is realized by channel-wise concatenate. This is used for extracting high resolution feature from the shallow layers. It can be seen that, during the process of encoding, the resolution of feature is decreased to quite low value, in order to absorb the more abstract features. However, quite low resolution impairs the accuracy of image segmentation. Thus, the skip connection is utilized to help extract rich features at low level. Low level information is also significant in IR tasks, thus this shortcuts is remained in this network architecture.

\subsection{Loss function}
Frobenious norm is used in our method, and the square of F-norm is used in training the first network, which is expressed as:
\begin{equation}
	L(\theta)=\ \frac{1}{2n}\sum_{x}\left|\left|R(\theta)-\ x\right|\right|^2
	\tag{5}
\end{equation}

This kind of loss function is widely used in computer vison to handle tasks related to images, for it has an advantage of fast speed compared with mean absolute error loss function. 

To fit our denoising task, based on the F-norm loss function, a penalty factor has been used in training process to pay specially attention on the false alarm point. The mathematical formula of the asymmetric loss function is shown below:

\begin{equation}
	L\left(\theta\right)=\frac{1}{2N}\sum_{i}{\left|\left|R\left(\theta\right)-\ x_i\right|\right|^2\times(\alpha+\beta\times x_i)}\ 
	\tag{6}
\end{equation}
In this loss function, y is the value from noise map, which is the value of 0 or 1 for clean or noise pixel. The penalty factor \((\alpha+\beta\times y)\) can be used to adjust the false alarm rate and the missing rate.

\section{Experiment Details}
Our method utilizes a forward deep neural network as the first stage, to convert all salt and pepper noise pixels to 0 value, while leaving clean pixels unchanged. In the second stage, a DRUnet is utilized to denoise image contaminated by purely black dots.

The original clean image and corrupted image are shown in Figure 2. Note that the salt and pepper noise level is 20\%, and the noise value is (16, 239), which is different from the common salt and pepper noise extreme value (0,255).  

\begin{figure}[htbp]
	\centering
	\subfigure[]{             
		\includegraphics[width=4cm]{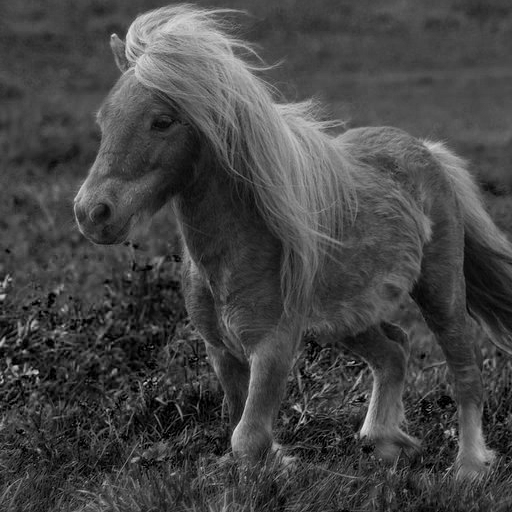}}
	% \hspace{0in}
	\subfigure[]{
		\includegraphics[width=4cm]{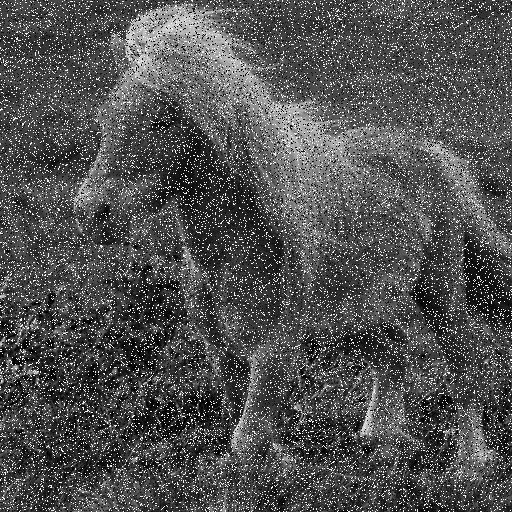}}
	\caption{Clean image, and 20\% corrupted image}
\end{figure}

\subsection{Detection of noise pixels}
In the first stage, a deep neural network is trained to detect the noise positions, called as Noise Map. On the noise map, 0 pixel value indicates the clean pixel, while 255 pixel value indicates the salt and pepper noise pixel. In the training step, the input of the model is a contaminated model and the ground truth is the noise map. Adam optimize is utilized, with batch size = 4. The starting learning rate = 1e-4, and decreased to 1/10 every 100 epochs. The loss figure of training is shown in Figure 3. After obtaining this noise map, it is possible to convert noise pixels to 0 on the original contaminated image.

The purpose of this training design is to shift all noise values to 0, but the problems is that some region with high values is mistaken for salt noises. Thus, a large area of black dots is observed in the processed image, as shown in  Figure 4. This is resulted by the high false alarm rate of the forward deep neural network. The large corrupted area can not be dealt well by the denoising model in the second stage.
\begin{figure}[htbp]
	\centering
	\includegraphics[width=7cm]{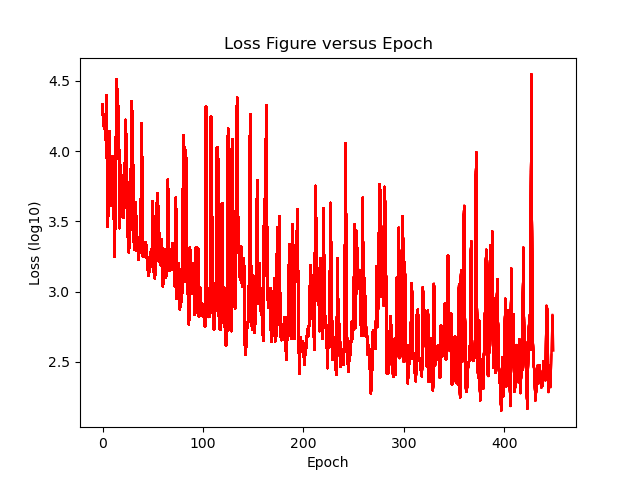}
	\caption{log10(Loss) versus epochs}
\end{figure}
\begin{figure}[htbp]
	\centering
	\includegraphics[width=4cm]{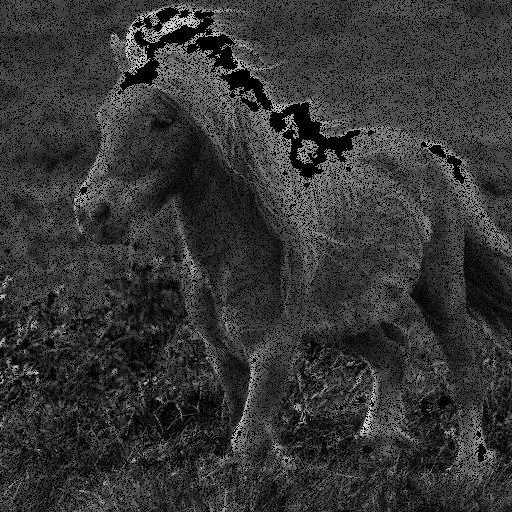}
	\caption{Large area mistaken for noises}
\end{figure}

Thus, a penalty factor is utilized in training to decrease false alarm rate. During the step of calculating loss,  the weights of clean pixel is 1.1 times larger than noise pixel. The loss function for the whole image has already been shown in previsou section in Equation 6. For a specific pixel, this is expressed as:
\begin{equation}
	L\left(\theta\right)=\left(R\left(\theta\right)-x\right)^2\cdot\left(\alpha+\beta\times x\right)
	\tag{7}
\end{equation}
\(L\left(\theta\right)\) is is the loss functions of this pixel, \(R\left(\theta\right)\) is the output pixel value, x is the value on the noise map. In other words, x can be either 1 or 0 for noise or clean pixel  respectively. Thus, the penalty factor \( (\alpha + \beta \times x)\) can adjust the false alarm rate and missing alarm rate, in which \(\beta\) has negative correlation with false alarm rate. 

The training set is the same as before, excpet the loss function. The loss figure is shown in Figure 5.
By this new loss function, the large area in the image is removed, and the model can detect the noises correctly, as shown in Figure 6.

% \begin{figure}[htb]
	% \centering
	% \includegraphics[width=4cm]{img/good-forward deep neural network.png}
	% \caption{Noise is detected correctly with new loss function }
	% \end{figure}
\begin{figure}[htbp]
	\centering
	\includegraphics[width=7cm]{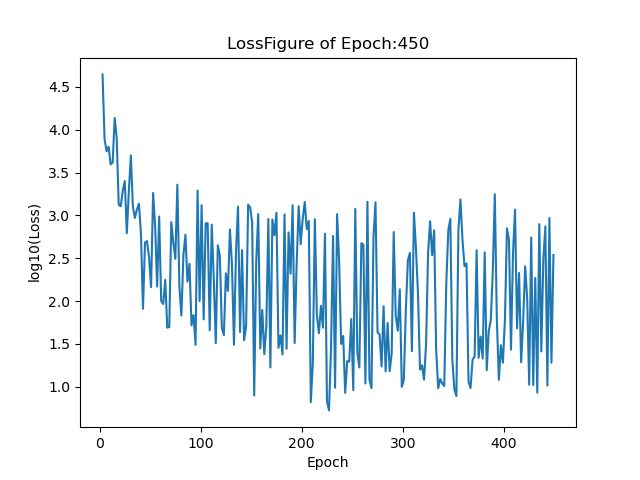}
	\caption{log10(Loss) versus epochs with new loss function}
\end{figure}
\begin{figure}[H]
	\centering
	\includegraphics[width=4cm]{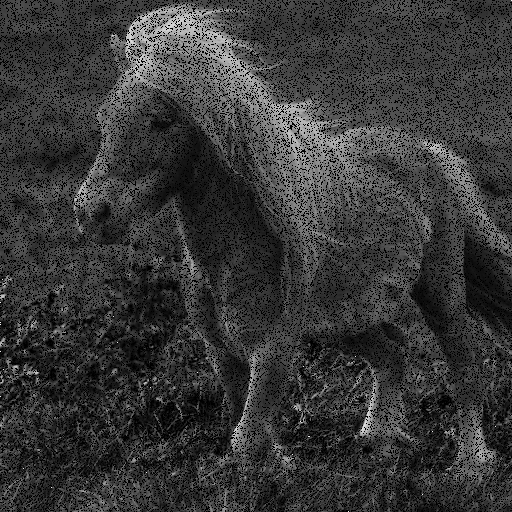}
	\caption{Noise is detected correctly with new loss function}
\end{figure}

\subsection{Remove the pepper noise}
The purpose of the second stage is to exclusively remove the 0 value noise pixels. During the training step, the input of DRUnet is an image contaminated by black dots with 0 value, and the ground truth is the clean image. The training parameters are similar as before. The resultant loss figure is shown in Figure 7.
\begin{figure}[htbp]
	\centering
	\includegraphics[width=7cm]{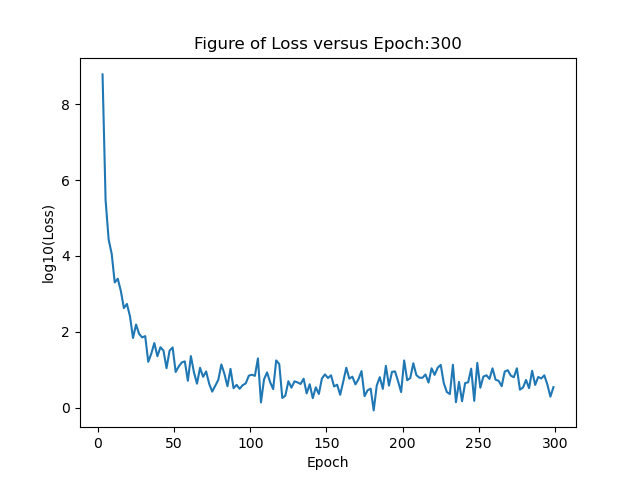}
	\caption{log10(Loss) versus epochs}
\end{figure}

Compared with the one-stage denoising method, which is also done by DRUnet, our proposed method achieves better performance. In the former case, the DRUnet learns to output a clean image from a corrupted image.
The comparison is shown in Figure 8. The direct denoising method achieves PSNR = 35.52dB, and our method achieves PSNR = 39.07dB, which is a significant improvement. In addition, our method can recovery more details, allowing the resultant image look more clear.

\begin{figure}[htbp]
	\centering
	\subfigure[One-stage, PSNR = 35.52dB]{             
		\includegraphics[width=4cm]{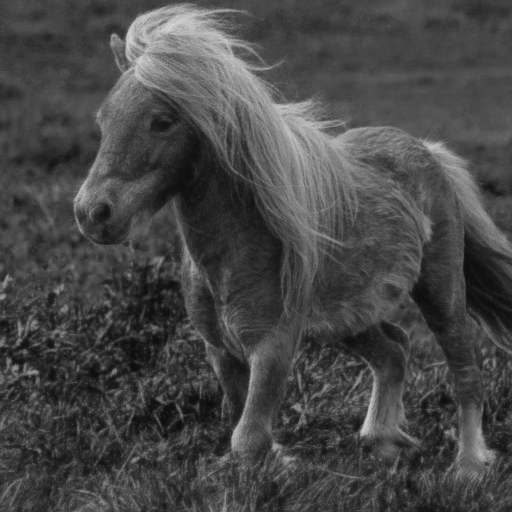}}
	% \hspace{0in}
	\subfigure[Ours, PSNR = 39.07dB]{
		\includegraphics[width=4cm]{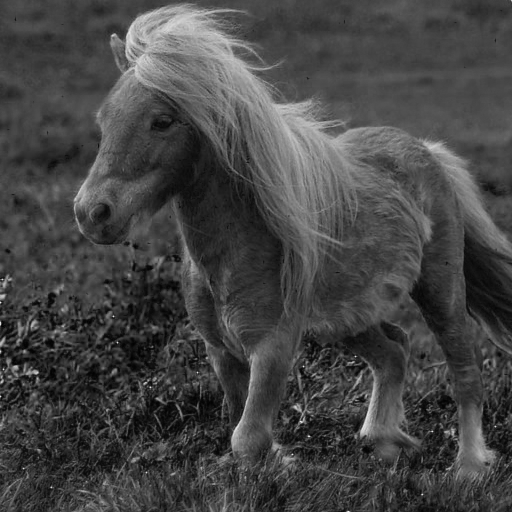}}
	\caption{Our method has improved performance quantitatively and visually}
\end{figure}

\section{Conclusion}
Consider that the salt and pepper noise in real application may not always be at extreme value (0,255), we have proposed a two-stage method to remove non-extreme value salt and pepper noise. The first stage converts all noise value to 0 by a forward deep neural network, and the second stage removes these processed noises by DRUnet. In the training step of forward deep neural network, a new penalty factor is inserted in the classical Frobenius norm to adjust the false alarm rate and missing alarm rate, removing large area of black dots. 
Our method has improved the performance quantitatively and visually, compared to directly remove the salt and pepper noise.
Our future work is to improve the performance of this method on high noise density contaminated cases.

	\end{document}